\documentclass[11pt]{article}
\usepackage{float}
\usepackage{morefloats}
\usepackage[final]{acl}
\usepackage{hyperref}
\usepackage{times}
\usepackage{latexsym}
\usepackage[T1]{fontenc}
\usepackage[utf8]{inputenc}
\usepackage{microtype}
\usepackage{inconsolata}
\usepackage{graphicx}
\usepackage{placeins}
\graphicspath{{figures/}}

\usepackage{placeins}
\usepackage{booktabs}
\usepackage{array}
\usepackage{multirow}
\usepackage{ragged2e}

\usepackage{enumitem}
\usepackage{float}
 \usepackage{amsmath, amssymb} 

\usepackage{tcolorbox}
\tcbuselibrary{skins, breakable}
\usepackage{arydshln}
\newcommand{\arr}[1]{\textcolor{black}{#1}}
\usetikzlibrary{calc,   arrows.meta}

\title{Is the ACL Responsible NLP Checklist a Box-Ticking Exercise? \\ A Large-Scale Analysis of EMNLP 2025}

\author{ 
    Nusrath Jinnath \ \ \ 
    Wei Zhao 
    \\[0.5em]
    The Aberdeen NLP Research Group\\
    University of Aberdeen\\[0.4em]
    Project website: \href{https://checklist.nlp4sci.com/}{https://checklist.nlp4sci.com/} \\[0.4em]
    \href{mailto:n.jinnath.22@abdn.ac.uk}{n.jinnath.22@abdn.ac.uk} \ \ 
    \href{mailto:wei.zhao@abdn.ac.uk}{wei.zhao@abdn.ac.uk}
  }

\begin{document}
\maketitle

\begin{abstract}
Responsible NLP practice includes a) transparency, b) ethics, and c) societal impacts. The Responsible NLP Checklist aims to push these goals, and promote responsible practice. Recently, ACL released the EMNLP 2025 Checklists to aid transparency on the current research practice, which we focus on. We curate and release the first two datasets of: a) all the checklist responses and justifications from the EMNLP 2025 Main and Finding tracks; b) checklist reference linking to paper sections.  We also provide the first analysis of recent EMNLP Checklists, by examining $73,922$ responses and justifications to them. For the Main track, we find that authors isolate ethics questions of the Checklist from the paper's bulk, mimicking the trend of ethics being an afterthought. We then examine \texttt{NO} responses. We find $44.9\%$ of justifications are poor or bad-faith, being brief or empty.  Then, we find significant issues with the checklist design and effort of authors, namely that $6\%$ of all checklists contained logical contradictions between parent and child responses. We also find evidence of surface compliance\footnote{Compliance refers to how often authors tick the box (answering YES in the checklist), meaning that they follow the ACL guidelines to report practices in the paper regarding datasets, ethics, computing, etc.} for responsible ethics, with $53\%$ authors dismissing potential risks or social impacts of their work, for which there should be none. We compare this to the Findings track, noticing a similar trend in both tracks. Lastly, we discuss the implications of the checklist design and provide recommendations for future checklist iterations. Including: a) enforcing a minimum word count, b) enforcing more scrutiny on the risks of appliances. 

\end{abstract}

\section{Introduction} 
\label{sec:intro}

\arr{Several} professional \arr{computing} bodies outline Codes of Conduct to nurture behaviour that will support the public good. Every body emphasises duty to the public, but differ on their guidance. Popular bodies include the British Computer Society (BCS), the  Association for Computing Machinery (ACM), and Institute of Electrical and Electronics Engineers (IEEE) \arr{Computing}. The ACM Code of Ethics~\cite{Anderson1992ACMCO} emphasises to ``avoid harm'' and ``contribute... benefit of society'' for its professionals, by minimising the threats to security and privacy and considering potential impacts on all stakeholders. The end goal is to maintain public trust, safety, and health. 

\begin{figure}[t]
    \centering
    \includegraphics[width=1.02\linewidth]{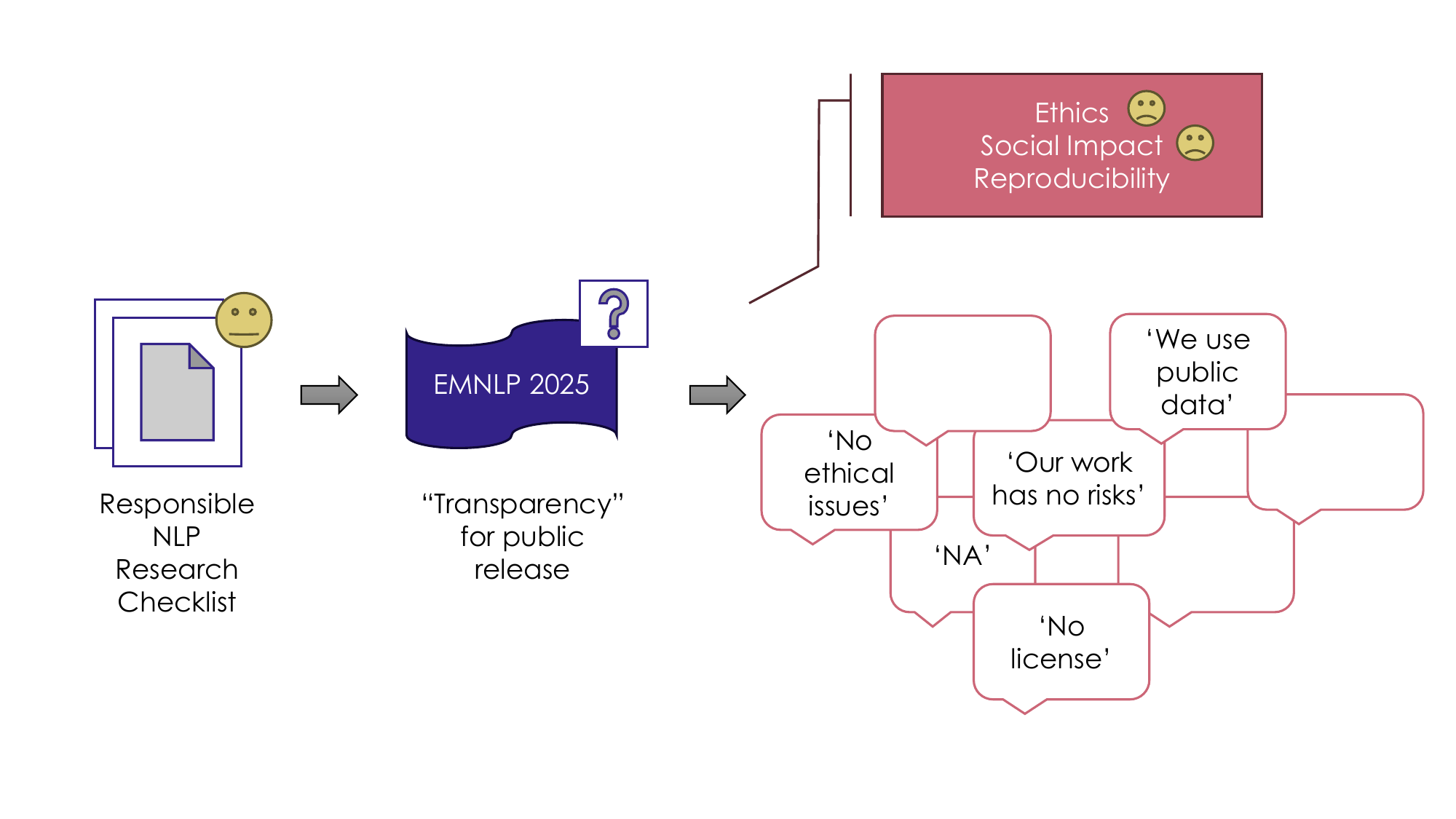}
    \caption{The ARR Responsible NLP Research Checklist has questions relating to ethics, social impact, and reproducibility. With EMNLP 2025 releasing said checklists publicly (for transparency), we made a pipeline to analyse the practices. We provide the first analysis with our own pipeline which created the Checkbox Corpus of 3,214 papers and checklists. In CheckBox, we crucially found $\frac{2}{3}$ are not addressed properly. We provide examples of (amended) the most common justifications found in the Corpus for \texttt{NO} justifications (including empty responses).}
    \label{fig:placeholder}
\end{figure}
In NLP, the wider \arr{ACL} body
adopts 
its own 
Code of Ethics to encourage responsible research practice.  
For NAACL 2021, 
\arr{paper} submissions \arr{are} required \arr{to fill out} the Reproducibility Checklist\footnote{\url{https://2021.naacl.org/calls/reproducibility-checklist/}}, based upon \citet{Dodge2019} and \citet{JMLR:v22:20-303}'s work. This \arr{checklist} focused \arr{on} reproducibility, and transparency for good scientific research. Additionally, following the Code, NAACL 2021 included an Ethics Review; asking questions relating to privacy, identifiable information or potential harms. All submissions must answer both, with authors encouraged to consider the ethical questions before submission. \arr{Given the importance of datasets for NLP systems}, \citet{Rogers2021}
introduced the Responsible Data Use Checklist to encourage safe and legal use of data. Their work aimed for authors to consider reproducibility, responsible data usage, and broader impacts more.

The ARR Responsible NLP Research~\citep{ACLRollingReview}  checklist was based \arr{on}
the latter checklists and the NeurIPS 2021 checklist~\cite{neurips_2021_program_chairs_beygelzimer_dauphin_liang_vaughan_2021}. This \arr{ARR} Checklist considers reproducibility, broader impacts on society (both negative and positive), and research ethics to promote responsible research practice. This was updated for the 2024 cycle, by Anna Rogers. See Figure \ref{app:checklist} for current checklist used. By design, the checklist is meant to prompt authors to reflect on whether their paper meets responsible research standards. Typically, this is not available publicly. However, last year, EMNLP 2025 released checklists for all accepted papers\footnote{\url{https://aclrollingreview.org/responsible-nlp-checklist-appendices}}.

Previous studies have only looked at checklist responses and answer distribution, or provided a tool for pre-submission. \citet{Magnusson2023} looked at acceptance rates and focused on reproducibility, whilst \citet{Galarnyk2025} focused on tokenisation and efficiency of the programs. Neither have looked at (i) specific checklist questions, (ii) checklist justifications, (iii) relationships between questions, and iv) the impact of checklist design on author responses. Researchers have not considered author justifications, nor the implications on best practice. Furthermore, their work concerned the previous checklist iterations, such as the Reproducibility Checklist. Henceforth, the impact of the Research Checklist has yet to be understood or analysed. 

The aim of this study was to evaluate the efficacy of the ARR Responsible NLP Research Checklist. With our pipeline, we have gathered 73,922 responses from EMNLP 2025.  We provide the first analysis of author's reasons towards these Checklist items.  
Our findings are outlined as follows: (i) 2 out of 3 best practices for responsible research are not being addressed consistently; (ii) Over half of authors do not consider potential risks of their work; (iii) Nearly half of authors do not \arr{adequately justify why they answered ``NO''};
(iv) Many authors carelessly fill out checklists, with a significant amount of \arr{logical errors}.
We repeat our analysis with findings, and find similar trends of ethical and societal questions of responsible research being less practised. \arr{Further}, we discuss the implications of the checklist design, author's responses, and provide recommendations for future checklist iterations. 

\section{Related Work}
\label{sec:related}

 \paragraph{ACL checklist analysis.} \citet{Magnusson2023} examined 10,405 anonymous submissions across across multiple tracks (e.g. EMNLP and NAACL 2021).  Their work provided the first quantitative analysis of the (previous) NLP Reproducibility Checklist; crucially, they found papers with more \texttt{YES} answers had higher acceptance rates. Further, they observed authors attempting to gamify the checklist. Some authors gave identical, bad faith, responses to every item. For example, all \texttt{YES}, or all \texttt{NO}. This suggests the checklist being taken less seriously than it should be. In contrast, \citet{Galarnyk2025} introduced ConfReady as a systems demonstration. Confready is a retrieval-augmented generation tool that helps authors draft checklist responses. Although they released a dataset of 1,975 ACL 2023 checklist responses, they did not analyse the responses. Instead, their analysis focused on token-level statistics and their tool is for presubmission, while our work analyses post-submission responses, and their justifications for EMNLP 2025, e.g., the adequacy of \texttt{NO} justifications.

\paragraph{Adjacent approaches to responsible practice.} Responsible practice is tackled by other angles, too. Tools like SciFact \citep{Wadden2020} and automated peer review systems \citep{Liu2023} evaluate the credibility of scientific claims. More recently,  \citet{Thomson2025} conducted a longitudinal survey of 149 NLP and ML researchers in 2022 and again in 2024, highlighting attitudes toward reproducibility were gradually improving, though significant barriers remained. In the same year, \citet{Karamolegkou2025} took a different approach with EthiCon: they extracted 1,580 ethics statements from ACL Anthology papers. With this, they compared concerns raised in the papers against a general-public survey.

\section{Responsible NLP Research Checklist}
\label{sec:checklist}

The Responsible NLP Research Checklist has 23 questions in total. Every question must be answered either: \texttt{YES}, \texttt{NO}, or \texttt{N/A}. It has 5 parent questions, and 18 subquestions. 
Each parent question has at least one subquestion. These subquestions always have the parent's letter in front. Only subquestions can be justified. The checklist taxonomy can be found in Table \ref{tab:responsible_nlp_taxonomy}.
 
\begin{table}
\footnotesize
\centering
\begin{tabular}{p{0.5cm}cp{1.7cm}p{3.9cm}}
\toprule
& ID & Subquestion & Description \\
\midrule

\multirow{2}{*}{\rotatebox[origin=c]{90}{\textbf{A. Mandatory}}}
& A1 & \tiny \textsc{Limitations}
& Describe the limitations of the work \\
\cmidrule{2-4}
& A2 & \tiny \textsc{Risks}
& Discuss potential risks of the work \\ \\

\midrule

\multirow{6}{*}[-14ex]{\rotatebox[origin=c]{90}{\textbf{B. Artifacts}}}
& B1 & \tiny \textsc{Citations}
& Cite the creators of scientific artifacts used \\
\cmidrule{2-4}
& B2 & \tiny \textsc{Licenses}
& Discuss licenses or terms of use/distribution for artifacts \\
\cmidrule{2-4}
& B3 & \tiny \textsc{Intended Use}
& Explain whether artifact use is consistent with its intended use and specify the intended use of created artifacts \\
\cmidrule{2-4}
& B4 & \tiny \textsc{Privacy}
& Discuss checks for personally identifiable or offensive content and anonymization steps \\
\cmidrule{2-4}
& B5 & \tiny \textsc{Documentation}
& Document artifact coverage (e.g., domains, languages, linguistic phenomena, demographics) \\
\cmidrule{2-4}
& B6 & \tiny \textsc{Statistics}
& Report artifact statistics (e.g., number of examples, train/dev/test splits) \\

\midrule

\multirow{4}{*}[-10ex]{\rotatebox[origin=c]{90}{\textbf{C. Experiments}}}
& C1 & \tiny \textsc{Resources}
& Report model size, computational budget, and computing infrastructure \\
\cmidrule{2-4}
& C2 & \tiny \textsc{Experiment Setup}
& Describe the experimental setup, hyperparameter search, and best hyperparameters \\
\cmidrule{2-4}
& C3 & \tiny \textsc{Statistics}
& Report descriptive statistics and specify whether results are from a single run, mean, maximum, etc. \\
\cmidrule{2-4}
& C4 & \tiny \textsc{Implementation}
& Report implementations, package versions, models, and parameter settings \\

\midrule

\multirow{5}{*}[-8ex]{\rotatebox[origin=c]{90}{\textbf{D. Human Annotators}}}
& D1 & \tiny \textsc{Instructions}
& Report the instructions given to participants \\
\cmidrule{2-4}
& D2 & \tiny \textsc{Recruitment}
& Describe participant recruitment, compensation, and payment adequacy \\
\cmidrule{2-4}
& D3 & \tiny \textsc{Consent}
& Explain how consent was obtained and how participant data are used \\
\cmidrule{2-4}
& D4 & \tiny \textsc{Ethics Approval}
& State whether the study received ethics approval \\
\cmidrule{2-4}
& D5 & \tiny \textsc{Demographics}
& Report demographic and geographic characteristics of participants \\

\midrule

\multirow{1}{*}[-1ex]{\rotatebox[origin=c]{90}{\textbf{E. AI}}}
& E1 & \tiny \textsc{AI Assistance}
& Disclose the use of AI assistants in research, coding, or writing \\

\bottomrule
\end{tabular}
\caption{Taxonomy of the Responsible NLP Checklist (ARR/EMNLP 2025).}
\label{tab:responsible_nlp_taxonomy}
\end{table}

\subsection{Justification Guidelines}
We set parents as \(P\), and child subquestions as \(S\). If $S=$ \texttt{N/A}, then no justification is required. Otherwise, these parents gatekeeps the child questions. 

\begin{description}

   \item[$P=$ \texttt{NO} \arr{\textsc{or}} \texttt{N/A}] Justifications are not required
   \begin{itemize}[leftmargin=0pt]
      \item Every $S=$ \texttt{NO} \arr{\textsc{or}} \texttt{N/A}. 
   \end{itemize}

   \item[$P=$ \texttt{YES}] Justifications required if $S=$ \texttt{YES} \arr{\textsc{or}} \texttt{NO}. 
   \begin{itemize}[leftmargin=0pt]
      \item $S=$ \texttt{YES} $\rightarrow$ Authors must cite the relevant section in the paper.  
 
      \item $S=$ \texttt{NO} $\rightarrow$ Authors must explain why the practice was not included (Box \ref{box:no-responses}) in the paper, through two ways: (i) why the practice was not included and (ii) provide the missing information that is not provided in the paper. 
   \end{itemize}
 
\end{description}




\section{The CheckBox Corpus}
\label{sec:methodology}
\arr{This section covers dataset construction (a pipeline for extracting both papers and checklists), as well as corpus statistics (\S\ref{sec:dataset}).}
 
\subsection{Dataset Construction}

\paragraph{PDF parser selection.}
  Choosing the best tool for accuracy was harder than expected. Academic papers are quite dense, and the formatting varies wildly between papers \arr{(e.g., heavy mathematical equations, and complex tables)}. Prioritising accuracy with such constraints is a difficult task.  We looked for tools that could preserve section ordering, hierarchy and contents as close as possible to the original PDF. Each tool was evaluated manually, through a sample of 20 papers. This sample was selected based on different observed structures (e.g. some were more mathematical, others more visual). We found PyMUPDF \citep{ArtifexSoftware2023} too shallow for academic papers, with section titles and content continuously mixing up. On the other hand, whilst Nougat \citep{Blecher2023} had better accuracy than PyMUPDF, it was much too slow. We pivoted, then, to consider Docling \citep{Auer2024} and GROBID \citep{Lopez2009}. Fortunately, these had 20\% higher accuracy, but this still did not preserve section ordering well. MinerU~\citep{MinerU2024} was briefly considered, but Marker~\citep{Paruchur2023} won since it maintained the best structure and highest accuracy over 30 documents. We note this can be less representative with the small sample size, and may cause issues in the paper-section linking algorithms. 

\paragraph{\arr{Our} pipeline.}
Thus, Marker is used to convert the PDFs of checklists and papers. The pipeline has three stages; in order: a) paper parsing, b) checklist extraction, c) and justification normalisation. For increased accuracy, steps b) and c) use \texttt{GPT 5.4-mini} constrained prompting. We set the temperature as 0, to minimise hallucinations. Public checklists for  industry, workshop, and system demonstration track were unavailable, so our pipeline covers only main and findings. All checklists and papers for main and findings were converted. 


\paragraph{Paper parsing.}
The default output of Marker is markdown, which uses hashtags for hierarchy. We found this did not preserve the original paper structure well, and therefore used JSON outputs instead. Further, we disabled image extraction, as our focus is paper content. Every \texttt{paper.json} has metadata and the page data. Of note here, the metadata includes the table of contents (TOC) for the paper. This TOC is made up of pages (page ids, titles, polygons). Every page contains child blocks of paper content. Taken together, this preserves section hierarchy well.

\paragraph{Checklist extraction.}
We used the same parameters (JSON outputs, no images) as the above section, for checklist extraction too. We believed checklists being smaller (2-3 pages) and semi-structured (23 questions) should give high or similar accuracy ratings, because these parameters were enough for paper extraction. To our surprise, this configuration gave about 75\% accuracy across 30 papers. We found two problems being either the checkbox symbols were misread, or there is missing justification text. 

And so, we had to pivot slightly. Marker's documentation\footnote{\url{https://github.com/datalab-to/marker}} noted using an LLM increases accuracy during the extraction process. We used constrained prompting to reduce hallucination. Even so, the issues still persisted: justification texts were missing or scattered throughout the JSON and the structure was not preserved. For example, the order is parent-answer, then child-answer-justification but the JSON would write parent-justification-child-answer-answer. 

Our next iteration involved Marker's \texttt{Structured Extraction Beta} model. This uses a JSON schema and an LLM to convert unstructured documents to structured ones. The structure, itself, is defined by the JSON schema. So, when parsing a PDF, the LLM must output a structured JSON that follows the JSON schema. Through this approach, the accuracy increased to 90\% across 15 papers. 

Our final pipeline looked like this: the checklist PDF is converted to markdown, by Marker. This PDF is also rasterised to JPEG. With these two, the LLM uses the markdown text for justification cleaning and the JPEG for checkbox symbols. Lastly, the LLM outputs a structured JSON following the checklist structure. See Figure \ref{fig:simple} for the diagram.

To validate this, we manually annotated a stratified sample of 71 papers ($2,911$ items) against the raw PDFs. We found only three errors of either checkbox misreads or cut-off justification text. Corresponding to 99.9\% item-level accuracy, \arr{we believe that our} overall results are unaffected \arr{by only a small portion of errors}.

\paragraph{Justification normalisation.}
With both checklists and papers being converted, we now look to bridge them together. When justifying \texttt{YES} responses in checklists, authors are called to reference sections in their paper. Every author does this differently, using styles such as: 
(i) formal naming (\emph{``Section 4.2 (Methodology)''}), 
(ii) numeric lists (\emph{``3, 4, 5''}), or 
(iii) shorthand (\emph{``App D, Sec.3''}).
Thus, without normalisation, linking checklist-paper sections is impossible. Our algorithm first uses regex extraction for numbered sections, appendices, and named sections. Then, for vague references, LLM fallback is used against the paper's TOC identifiers. The final stage validates all outputs against the paper's TOC, to mitigate hallucinations.

This pipeline achieved a 90.9\% section-text retrieval rate, for the entire corpus of 3,214 papers, and their corresponding 3,214 checklists enabling checklist references to be linked to paper section text for the first time.

\subsection{Corpus Statistics}
\label{sec:dataset}

Our data includes, for a given submission, the checklist responses ($\texttt{YES}, \texttt{NO}, \texttt{N/A}$ for each item), author justifications, and (normalised) section references to the paper. The paper title, link, and checklist link are also included. To our knowledge, when submitting to EMNLP 2025, authors were aware their responses will be made public to help transparency. 
\begin{table}
\centering
\footnotesize
\begin{tabular}{@{}lrr@{}}
\toprule
                          & \textbf{Main}   & \textbf{Findings} \\ \midrule
Parsed Responses          & 41,607 & 32,315   \\
\texttt{YES} (\%)                  & 57.8   & 55.49     \\
\texttt{NO} (\%)                   & 11.3   & 12.1     \\
\texttt{N/A} (\%)                  & 30.9   & 32.4     \\
\bottomrule
\end{tabular}
\caption{Corpus overview, for all 23 checklist questions.}
\label{tab:corpus-overview}
\end{table} 
 Across all items and conferences in Table~\ref{tab:corpus-overview}, all 3,124 papers and checklists are included. The analysis focuses on the checklists, mostly. 
 
\section{Main Track Results}
\label{sec:results}
\arr{This section covers the proportions of \texttt{YES}, \texttt{NO}, and \texttt{N/A} responses (\S\ref{sec:dist}), the quality of \texttt{YES} (\S\ref{sec:yes}) and \texttt{NO} (\S\ref{sec:quality}) justifications, logical inconsistencies in responses (\S\ref{sec:logic}), and a comparison with the findings track (\S\ref{sec:findings-comparison}).}


\subsection{Distribution of Checklist Responses}
\label{sec:dist}
\begin{figure}[htbp]
    \centering
    \includegraphics[width=\columnwidth]{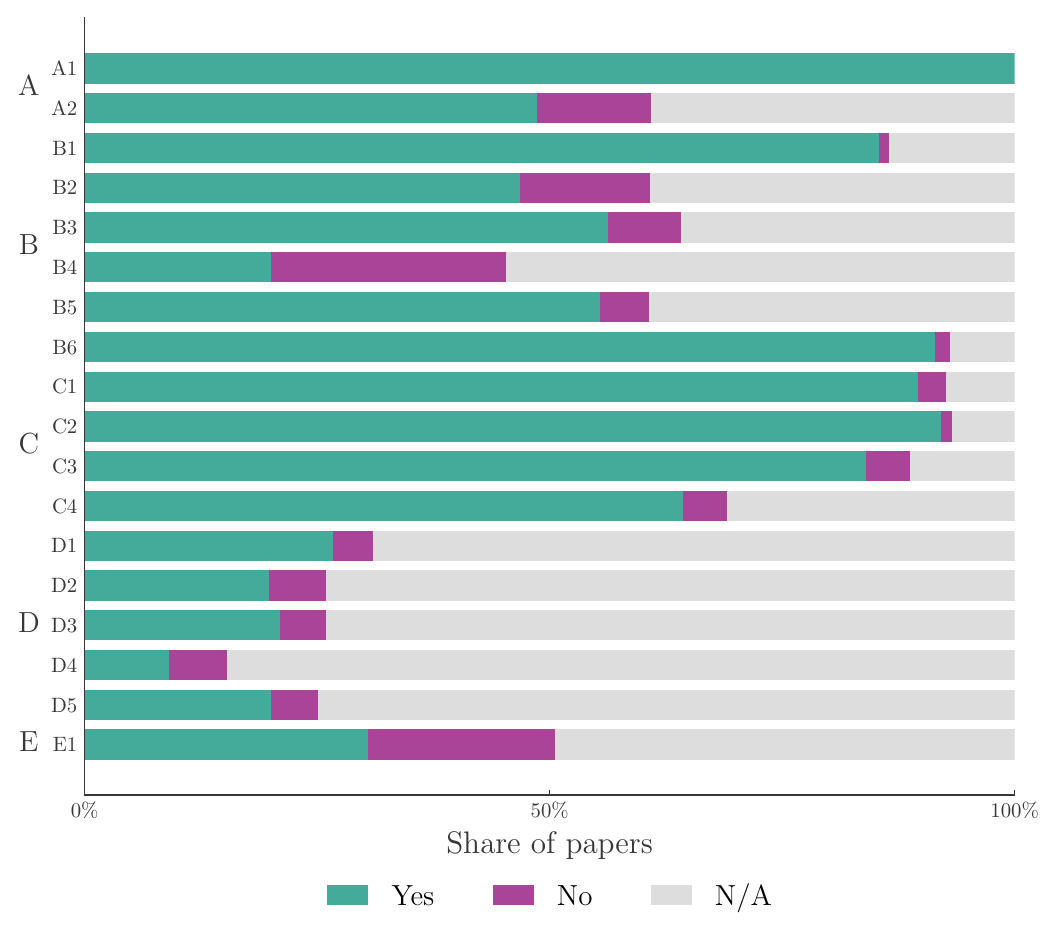}
    \caption{Checklist subquestion answer distribution across 1,809 main-track checklists}
    \label{fig:overview}
\end{figure}

\paragraph{Reproducibility questions have higher compliance than ethics and societal impact questions.} In Figure~\ref{fig:overview}, $A1$ and Parts $B$ and $C$ dominate the compliance ratings, with $A2$ and Part $D$ on the lower ends. This is the first instance of reproducibility being more compliant than ethics or societal impact. Specifically, $D$ is over 3.5x less compliant than the rest (including $E$ which concerns AI usage).  
 
\paragraph{Most authors dismiss risks and impacts of their application.} We focus on $A2$, ```Did you discuss any potential risks of your work?''' in the same Figure~\ref{fig:overview}. Examples of risks include (but not limited to) dual use, bias, or surveillance. We find over half of authors disregard this key responsible practice question. An important issue from this data is the violation of the ACM Code of Ethics. 

By ignoring potential risks or impacts of their application,  this inadvertently goes against public duty and NLP for social good. This is further supported by guidelines emphasising reflection on future use, but as the form does not incentivise such reflection, authors choose to skip it. 
  

\subsection{\texttt{YES} Justification Quality} 
\label{sec:yes}

We now examine the quality of the 'Yes' responses for each subquestion. We test this with a concentration score, calculated as: \( \tfrac{\text{Total Section References}}{\text{Unique Sections Cited}} \).

\begin{figure}[htbp]
\centering
\includegraphics[width=\columnwidth]{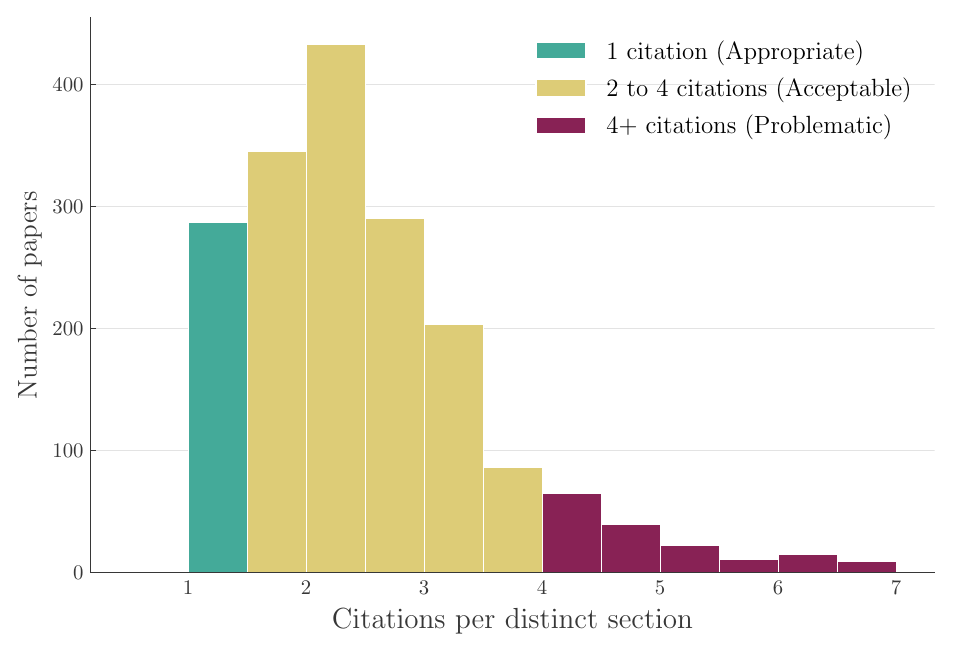}
\caption{Citation concentration scores distribution.}
\label{fig:main-cite-concentrate}
\end{figure}

\paragraph{Checklist questions are not specifically addressed in paper.} \label{sec:concentration}

Figure~\ref{fig:main-cite-concentrate} illustrates most checklists have a $Score $$< 4$; median concentration: 2.25, and the mean is 2.42. For example, an average author would reply  ``Section 3.1, 3.2'' (2 sections) to 1 question.
We classified checklists as ``problematic'' if they cited too many sections per question. 132 checklists have been labelled as such, having $Score>4$. For example, the worst problematic checklist contained $>40$ references pointing towards $<5$ paper sections. As the checklist aims to ensure best practices are followed when creating or using artefacts, such problematic checklists make verification much harder. \arr{For a rigorous check, a reviewer may have to verify whether the practices were actually followed by going through multiple paper sections for one question, which seems unlikely in practice due to time reasons. However, we note that doing so is necessary given the current checklist form, because the cited section sometimes does not actually address the checklist question. For example, we find papers citing the ``Limitations'' section for A2 (potential risks) sometimes only discuss technical details} instead of considering other stakeholders. 

Despite this, the current checklist submission form provides no automated feedback on referencing patterns, nor does it prompt authors to review their answers before submission. As such, there is nothing to flag this problematic behaviour which makes verification harder.

 \begin{figure}[htbp]
    \centering
    \includegraphics[width=\columnwidth]{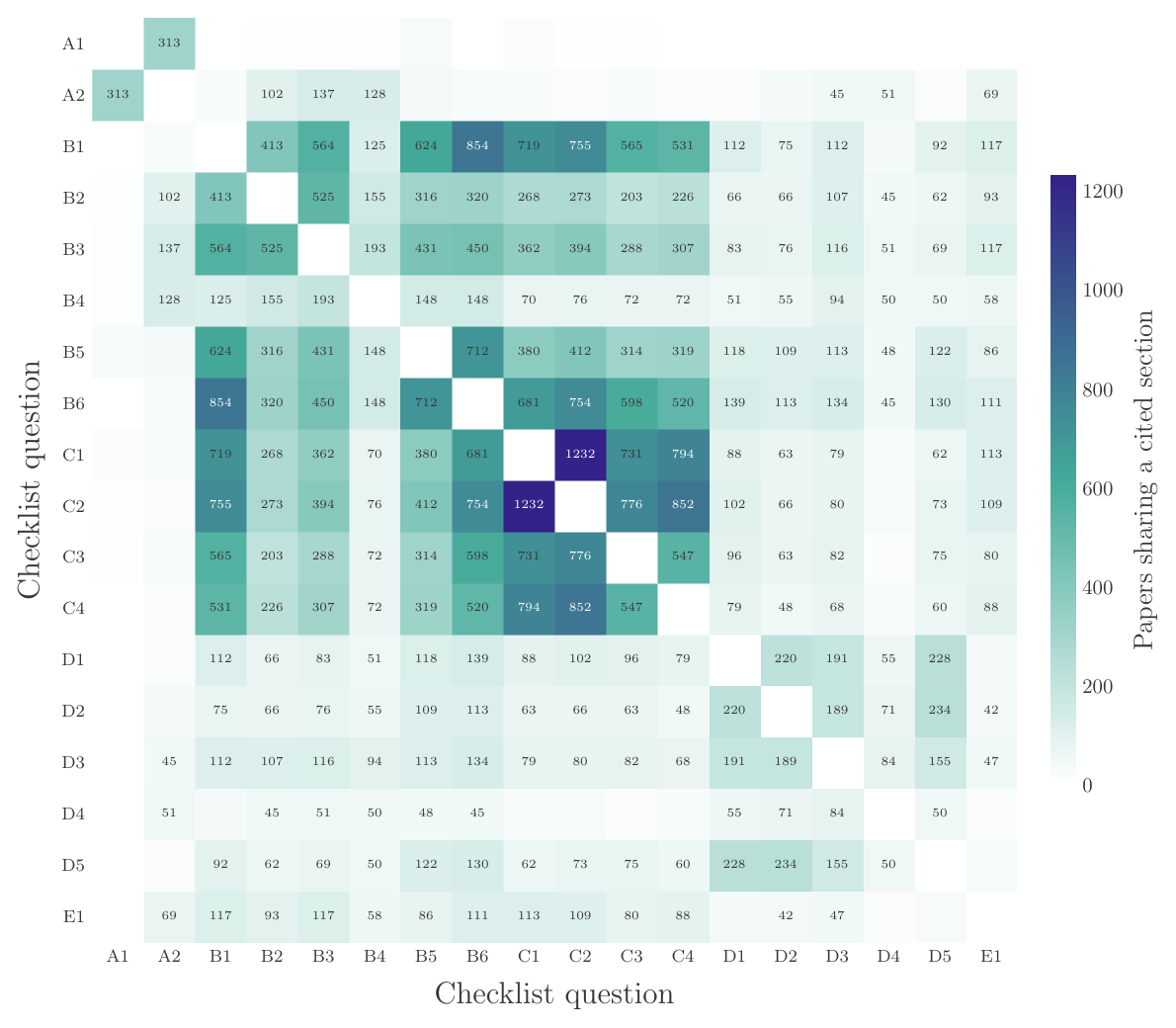}
    \caption{Co-citation matrix. Each cell records how many papers cite the same section for both questions. 
    }
\label{fig:main-cocite-matrix}
\end{figure}

We then examined which question pairs reference the same section. Figure \ref{fig:main-cocite-matrix} illustrates denser (more repetitions) with darker colours. Within checklist components are related, whilst cross checklist components are not related. 

\paragraph{References  are repeated for related and non related questions.} As expected, the densest areas are within checklist components (e.g. $B2$-$B3$ or $C1$-$C2$) in parts $C$ and $B$. 

Unexpectedly, the data is densest between $B1$ to $C4$, indicating large overlaps between references.  An issue arises from this. $B1$ asking ``did you cite the creators of your artefacts?'' and $C2$ asking ``did you describe the experimental setup?'' targets two separate practices for authors, but authors reference the same paper section. So in practice, \arr{the checklist has become a lazy exercise of section pointers, making it difficult to assess whether checklist questions are adequately addressed}.

Turning to the bottom, Part $D$ (human ethics, annotators) has clustered references within $D$. As such, Part $D$ exists isolated from Parts $B$ and $C$; cutting it off from the rest of the paper. $B1$-$C4$ mimics this isolation trend. $B4$ (data ethics) has the least pairings between any items in these nine items, even though it is a part of $B$ and \arr{arguably one of the most important ethical concerns in NLP.} Since $B4$ covers data anonymisation and privacy, one reason for this isolation could be that most experiments do not handle private data.
This is evidenced in Figure~\ref{fig:overview}, with $B4$ having the least compliance in B. Perhaps researchers may feel that data ethics does not belong in the main technical discussion as they are using public datasets. Consequently, $B4$ may benefit from being included in a separate component dedicated to data ethics as the current checklist might be outdated. Alternatively, another reason is the checklist not promoting reflection on data ethics enough, because $B4$ also has the highest '\texttt{N/A}' rate of all checklist components. The implication is most researchers either have data that does not need to be anonymised, or that researchers did not verify if the data complied with ethical standards. The implications of the latter is worse since data ethics is a main problem, which could tie into A2 (risks) as well.

\subsection{\texttt{NO} Justification Quality}
\label{sec:quality}

We evaluated the justification quality of 2,074 \texttt{NO} justifications (where $P$=\texttt{YES}) by word count. Justifications were grouped in either adequate (>10 words), brief (1-10), and empty (0) categories (see Table~\ref{tab:no-justification-examples} for examples). We used \(Failure = Empty + Brief\).


\begin{table}[h]
\centering
\small
\begin{tabular}{@{} p{0.18\columnwidth} 
                   >{\RaggedRight\arraybackslash}p{0.70\columnwidth} @{}}
\toprule
\textbf{Category} & \textbf{Example} \\
\midrule
\multirow{2}{*}{Adequate}
   & ``We don't release or use datasets or models, so licensing does not apply to our contribution.'' \\
\midrule
Brief & ``Not applicable to our work.'' \\
\midrule
Empty & (no text provided) \\
\bottomrule
\end{tabular}
\caption{Examples of \texttt{NO} justifications from the corpus, classified by word count.}
\label{tab:no-justification-examples}
\end{table} 

\begin{figure}[h]
    \centering
    \includegraphics[width=\columnwidth]{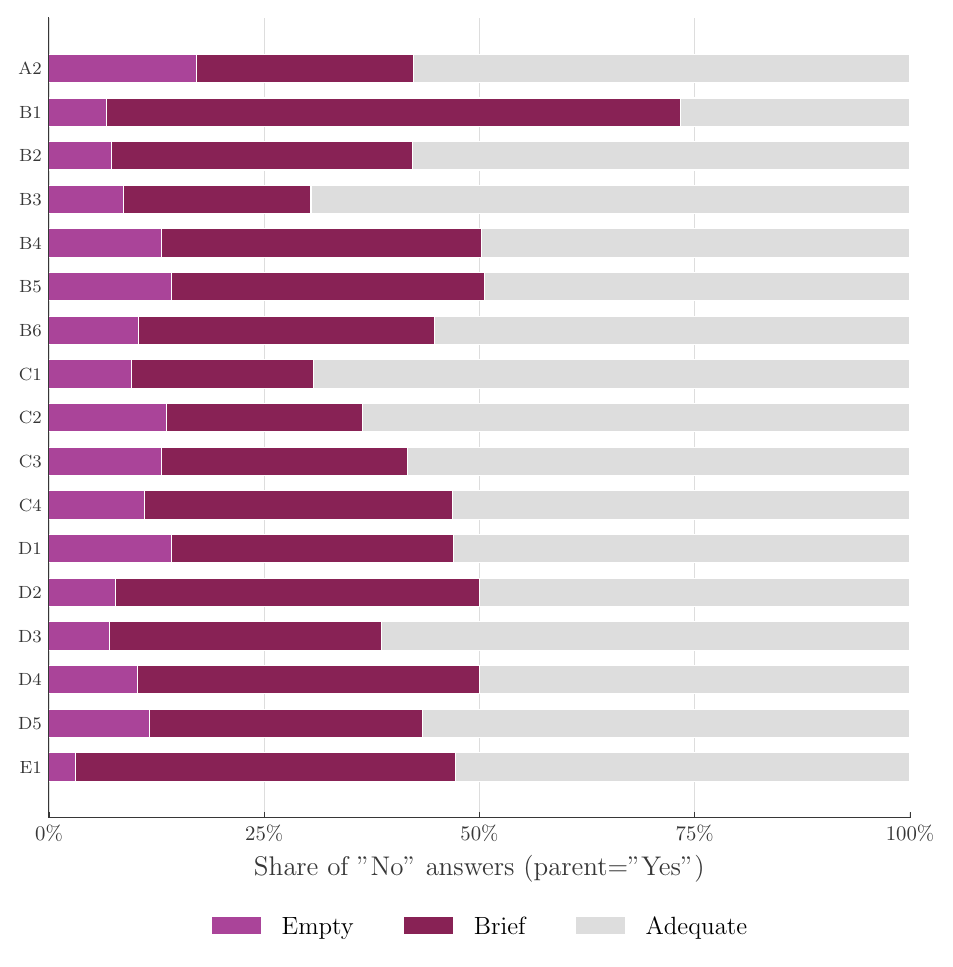}
    \caption{Justification quality for $S$=\texttt{YES} where the $P$ = \texttt{YES}, grouped by empty, brief (<10 words) and adequate (>10 words).}
    \label{fig:main-failures}
\end{figure}

\paragraph{Nearly half of all \texttt{NO} justifications are inadequate.} 44.9\% of \texttt{NO} justifications fall under brief or empty words (see Figure \ref{fig:main-failures}). 721 cases were brief, whilst 213 were empty. Hence, it can be reasoned that \texttt{NO} justifications fail on their primary function: accountability, because significant authors failed to meet even a minimal standard of explanation.  This is quite poor, especially as \texttt{NO} justifications are designed to encourage reflection throughout the research, on top of accountability near the end.

\paragraph{Lack of widespread effort for \texttt{NO} justifications.} We found $C$ had the least failure rate \arr{overall}. Two reasons could explain this: computational experiments ($C$) is heavily documented, and $C$ had the highest compliance. Hence, authors have more material to draw from when declining a subquestion. Briefly, we point back to the answer distribution in Figure \ref{fig:overview} and highlight $B$ and $D$. As illustrated, $B$ has a $>3×$  higher answer rate than D. Even then, surprisingly, $B$ and $D$ were between them even with the large disparity in answering rates. This challenges answer distribution being a factor in adequate author responses, as previously we assumed a paper with more $YES$ answers promotes better justifications. Instead, this raises a more troubling link between author's efforts and checklist justifications. With a lack of guardrails or clear incentives, authors do the bare minimum to be accepted instead of best practices.  This is evidenced by all subquestions suffer from  <75\% adequacy, instead of clustering around specific questions, indicating it to be pervasive and widespread.

\paragraph{One in six authors who claim their work carries no risks leaves the justification completely empty.} $A2$ has the highest empty rate in the corpus (17.1\%). Pairing with $A2$ having a high \texttt{NO} rate (see Figure~\ref{fig:overview}), this indicates a startling lack of following best practices as $A2$ is mandatory for all authors to fill out. Authors who did not follow best practices did not provide adequate explanations, which discounts formatting confusion.  By withholding an explanation \arr{regarding no risks}, authors bypass a critical safety check. It is vital to point out potential risks of new technologies in the earlier stages, so that mitigation strategies can be more developed. As such, safer and more ethical systems can be used for public good. Thus, \texttt{NO} justifications for $A2$ should be reflected on the most but the data indicates the opposite.

\paragraph{More authors justify using AI in research as opposed to critically considering potential risks of the system.} Unlike $A2$, $E1$ is not mandatory and it is newer: only added in recent years, compared to $A2$ which has been there near the start. Even so, briefly flipping to Figure \ref{fig:overview} we can see the \texttt{YES} answers are steadily catching up to $A2$. We also found $E1$ had more cases for $NO$ justifications with 324 cases vs $A2$ having 223 cases. Comparing the justification quality of \texttt{NO} answers, $E1$ has a significantly smaller empty rate of 3.1\%. Compared $A1$, $E1$ has nearly double the amount of brief responses (44.1\%) highlighting authors fill in the checklist with explanations of their use of AI, rather than risks associated with the NLP systems. In society where AI Usage in itself is debated, and could carry potential risks, it's interesting to note people defend their usage more than their actual work's impact.

\subsection{Logical Contradictions}
\label{sec:logic}
\begin{figure}[htbp]
    \centering
    \includegraphics[width=0.9\linewidth]{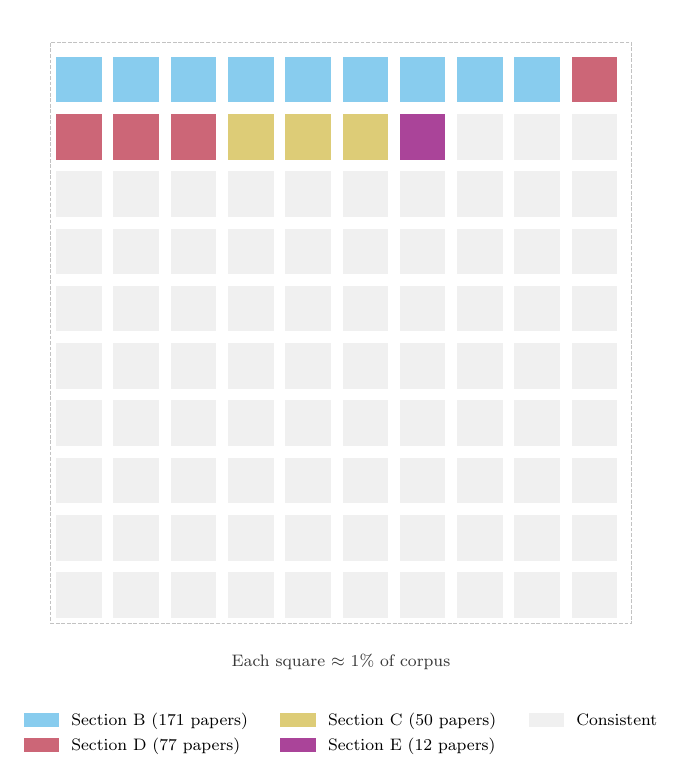}
    \caption{Inconsistency rate, wherein P=NO but the child question, S, is YES}
    \label{fig:main-waffle}
\end{figure}

According to the guidelines, if $P=$ \texttt{NO} \arr{\textsc{or}} \texttt{N/A}, then all $S=$ \texttt{NO} \arr{\textsc{or}}  \texttt{N/A}. We expected 0 contradictions.  We found $281$ papers (15.5\%) contain at least one case (see Figure~\ref{fig:main-waffle}). $B$ dominates the category at a substantial 9.5\% inconsistent rate.

\paragraph{Authors carelessly fill out answers.} Child questions should not be applicable if the parent is null. For example, $B$ ``Did you use or create scientific artifacts'' is \texttt{NO}, then it is impossible for $B5$ ```Did you provide documentation of the artifacts'' to be a \texttt{YES}. Yet, the data demonstrates that very contradiction occurs significantly. \arr{This problem may slow down human reviewers when reviewing a checklist, and potentially causes errors for} checklist data \arr{analysis.} Many parent questions clearly relate to the subquestions in a generalised manner. If an author marks a $P=$ \texttt{NO}, but a $S=$ \texttt{YES}, then the author should do the diligence to double check if the $P$ applies after. As, with the checklist's structure, it must apply. 

\paragraph{Vulnerabilities in reviewing.}
Another impact of these logical contradictions is on reviewer's. Negative answers on $P$ are signals to ignore the $S$ subquestions, but contradictions in the checklist slows this process down.   For example, if $D$ = \texttt{NO}, then  a reviewer is less likely to scrutinise $D4$ (ethics approval) if $D4$=\texttt{YES}. Which could cause errors down the line, since the later something is caught, the more of a hassle it is to fix.

\subsection{Findings Track Comparison} 
\label{sec:findings-comparison}
 We reran the full analysis on the 1,405 findings-track checklists to compare our results.

\begin{figure}[htbp]
    \centering
    \includegraphics[width=\columnwidth]{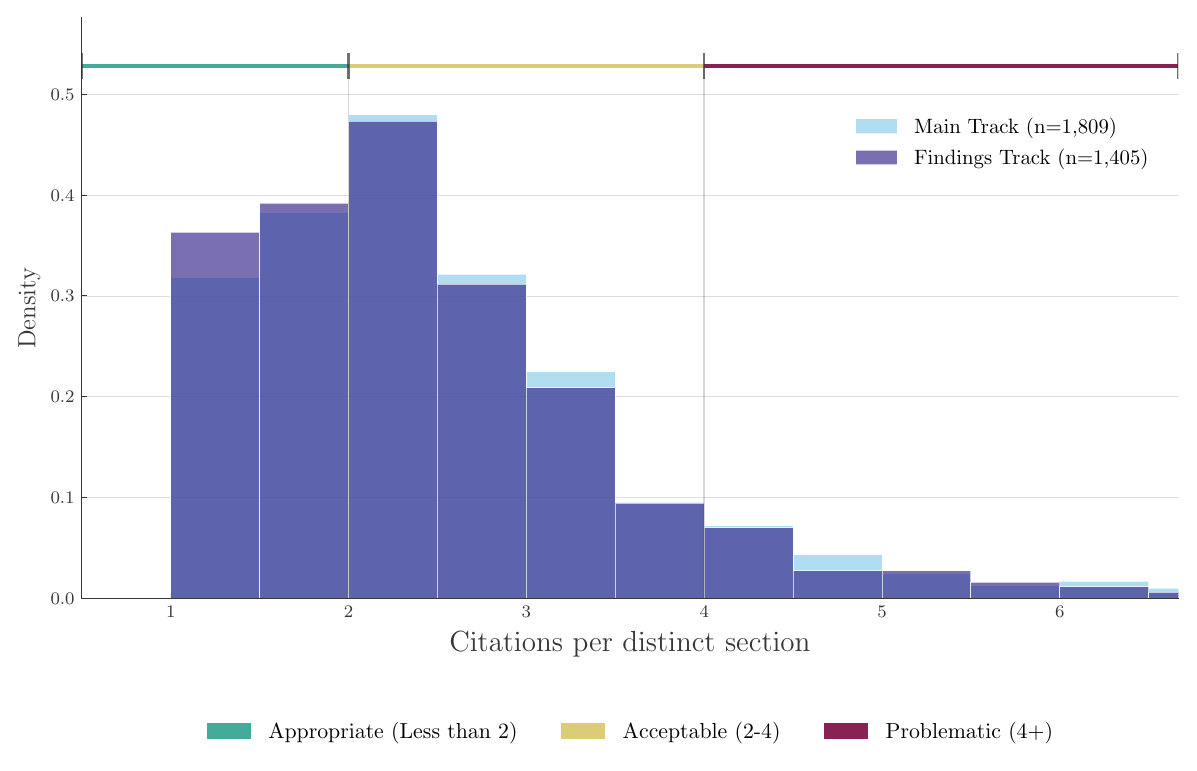}
    \caption{Citation concentration overlay for both tracks.}
    \label{fig:comp-concentration}
\end{figure}

\paragraph{Findings submissions have less overall compliance.} In general, reviewers route papers to the findings track when the core contribution is narrower or the evaluation is less exhaustive. Between the two tracks, findings has more \texttt{NO} and \texttt{N/A} for most questions (see Figure \ref{fig:comp-answers}) showcasing less overall compliance. Importantly, $B5$, $C4$ and $C1$ had the highest percentage differences between main and findings, with findings having less \texttt{YES} responses. This could indicate main having overall better quality in experimental setup or design, showcasing the checklist being used well for paper quality. However, the percentage difference is only $2\%$ which raises a question on how significant the checklist is at examining paper quality. Logically, since the main track has a lower acceptance rate (than the finding), the checklist responses should be more thorough.
When comparing the quality of justifications (see Figure \ref{fig:comp-inconsistency}), we find the previous questions ($B5$, $C4$ and $C1$) having better adequacy ratings when $P$=\texttt{YES} and $S$=\texttt{NO}. One implication is this higher adequacy compensates for the percentage differences between questions. Thus, this could imply the checklist does not examine paper quality as well as it should.

\paragraph{Checklist design encourages heavy concentration, and inconsistencies.} Figure~\ref{fig:comp-concentration} illustrating the concentration scores are very similar between tracks,  with findings having slightly higher outliers. To add on, both have similar medians of ~2 (when rounded), with extremes of $Score>8$, demonstrating  heavy concentration as a product of the checklist design.  This indicates authors map the checklist on their existing structure, regardless of acceptance track. Another common trend is logical inconsistencies (see Figure \ref{fig:comp-inconsistency}), with the overall inconsistency rate being higher at $18.1\%$ Taken together, this further supports the checklist design being unsuitable for distinguishing paper quality, as the issues persist between tracks but are not significantly worse in findings. 

\paragraph{Findings justifies ethical questions significantly less adequately.}   $A2$'s adequacy rating between tracks echoes findings echoing the main track issues. Even less authors (-7\% difference) adequately justify \texttt{NO} responses, with similar trends of authors adequately justifying $E1$'s (5.2\%) AI Usage more. One reason for the differences could be findings have 1,405 papers (less than 1809) which could make the percentage differences more noticeable between tracks. For example, findings has a lower overall failure percentage (44.6\%) but individual questions suggest the opposite. 

As such, these suggest the Checklist neither encourages best practices nor indicates paper quality, as it is not being used for its intended purpose during and before the research. 

\section{Discussion} 
\label{sec:discussion}
A key, repetitive theme emerging is questions concerning ethical practice or wider societal impact are insufficiently addressed or not attempted.
Of significance is $A2$ (risks), and $B4$ (data ethics) which are crucial in modern society. By treating risk considerations ($A2$) as optional, and providing justifications that are superficial (vague, short, or just a rehash of the question ``Our work contains no risks'') undermines the entire value of this question and ethical practice. Previous literature has made it clear advanced language models carry well-documented capacities for harm \citep{kumar-etal-2023-language}. For example, many models can contribute to bias amplifications \citep{ZADID2026100773, shah-etal-2020-predictive, DBLP:journals/corr/BlodgettO17, pmlr-v81-buolamwini18a}.
With authors not considering the impacts on society, there are less mitigations, and therefore it could lead to significant harms to public duty and public welfare. $B4$ (data ethics) exacerbates this further, as \citet{leins-etal-2020-give} found most NLP systems carry risks of either dual use or data ethics, and the guidelines recommend looking at the paper. However, we know that this had the worst compliance metrics. Every model will carry inherent risks from the previous model, and these will propagate, which significantly derails responsible practice. 

Our immediate recommendations are simple: 
(i) When declining a parent component, all child questions should be disabled. This will prevent logical contradictions, and force authors to rethink their answers;
(ii) Every justification for non compliance must not be empty;
(iii) A2 in particular should be given additional consideration by reviewers, beyond current standards. If risks are not discussed in the main content, reviewers should apply higher levels of scrutiny to answers in the checklist.

\section{Conclusion}
\label{sec:conclusion}

The main goal of the current study was to determine current rates of responsible practice in EMNLP 2025. This is the first study which has examined the associations between checklist adequacy, and responses. By providing a new pipeline to extract and analyse responses, this work offers a novel understanding of checklist compliance. The most obvious finding to emerge from this study was 53\% of authors dismissing potential risks or social impacts of their work. The second major finding was that 44.9\% of \texttt{NO} justifications being either brief or entirely empty, which signals a larger issue with current checklist formatting. These findings will be of interest to checklist designers and researchers primarily. Taken together, these results suggest that responsible research practice is not being promoted adequately. As such, we propose changes to help improve checklist response rates and help promote responsible research practice.  
This is also the first study to release a dataset of checklist responses linking to the cited sections, for future work in responsible practice analysis. 

\section*{Limitations}

These results warrant further investigation across conferences and years. This work is needed to confirm whether checklist reformations are needed for different venues, or if they are just characteristic of the ACL community.

Secondly, our analysis focuses on post submission checklists. In particular, we assumed that each published checklist is finalised. In practice, though, many papers undergo revision between checklist submission and camera-ready. This can lead to mismatches between the checklist responses, and the published text. As we do not have a way to detect this, we cannot quantify it either. This can affect compliance rates or other derived figures.  

Next, our \texttt{NO} justification quality analysis relies on word count. This is a more basic version compared to analysing the actual words. For example, a ten word justification could be perfectly adequate but our software would classify it as inadequate. Future work would include manual annotation by checklist researchers, to confirm the quality of justifications. 

Another area of future work is \texttt{YES} justification quality. Our second dataset includes referenced section texts for each question, but we did not have time to analyse it. Even so, this dataset can be used to examine the percentage of referenced sections that address the Checklist. For example, through manual validation, which is key to assess the quality of compliance at a deeper level.


\section*{Ethical Considerations}
Our dataset is based on publicly available EMNLP 2025 papers and their Responsible NLP Checklists. Although the dataset does not contain author names, it includes paper titles, abstracts, and content, which would allow for identifying author names. However, since this information is already publicly available in open-access publications, the privacy risk is limited. We also note that our dataset does not contain sensitive personal information beyond standard publication metadata, and is distributed solely for research purposes, following the licensing terms of the ACL Anthology. Our goal is to understand current responsible research practices; therefore, we report only overall results and do not evaluate or compare the checklist quality of individual papers, authors, or institutions.

\bibliography{custom}
\FloatBarrier

\clearpage
\appendix

\section{The Responsible NLP Checklist}
\label{app:checklist}

\begin{tcolorbox}[colback=white, colframe=black, title=Responsible NLP Checklist Questions, breakable]

\subsection*{A. Questions mandatory for all submissions.}
\label{check:A}
\begin{enumerate}
    \item[A1.] \label{check:A1} Did you describe the limitations of your work?
    \item[A2.] \label{check:A2} Did you discuss any potential risks of your work?
\end{enumerate}

\subsection*{B. Did you use or create scientific artifacts? (e.g. code, datasets, models)}
\label{check:B}
\begin{enumerate}
    \item[B1.] \label{check:$B1$} Did you cite the creators of artifacts you used?
    \item[B2.] \label{check:B2} Did you discuss the license or terms for use and/or distribution of any artifacts?
    \item[B3.] \label{check:B3} Did you discuss if your use of existing artifact(s) was consistent with their intended use, provided that it was specified? For the artifacts you create, do you specify intended use and whether that is compatible with the original access conditions (in particular derivatives of data accessed for research purposes should not be used outside of research contexts)?
    \item[B4.] \label{check:$B4$} Did you discuss the steps taken to check whether the data that was collected/used contains any information that names or uniquely identifies individual people or offensive content and the steps taken to protect/anonymize it?
    \item[B5.] \label{check:B5} Did you provide documentation of the artifacts, e.g., coverage of domains, languages, and linguistic phenomena, demographic groups represented, etc.?
    \item[B6.] \label{check:B6} Did you report relevant statistics like the number of examples, details of train/test/dev splits, etc. for the data that you used/created?
\end{enumerate}

\subsection*{C. Did you run computational experiments?}
\label{check:C}
\begin{enumerate}
    \item[C1.] \label{check:C1} Did you report the number of parameters in the models used, the total computational budget (e.g., GPU hours), and computing infrastructure used?
    \item[C2.] \label{check:C2} Did you discuss the experimental setup, including hyperparameter search and best-found hyperparameter values?
    \item[C3.] \label{check:C3} Did you report descriptive statistics about your results (e.g., error bars around results, summary statistics from sets of experiments), and is it transparent whether you are reporting the max, mean, etc. or just a single run?
    \item[C4.] \label{check:C4} If you used existing packages (e.g., for preprocessing, for normalization, or for evaluation, such as NLTK, SpaCy, ROUGE, etc.), did you report the implementation, model, and parameter settings used?
\end{enumerate}

\subsection*{D. Did you use human annotators (e.g., crowdworkers) or research with human subjects?}
\label{check:D}
\begin{enumerate}
    \item[D1.] \label{check:D1} Did you report the full text of instructions given to participants, including e.g., screenshots, disclaimers of any risks to participants or annotators, etc.?
    \item[D2.] \label{check:D2} Did you report information about how you recruited (e.g., crowdsourcing platform, students) and paid participants, and discuss if such payment is adequate given the participants' demographic (e.g., country of residence)?
    \item[D3.] \label{check:D3} Did you discuss whether and how consent was obtained from people whose data you're using/curating (e.g., did your instructions explain how the data would be used)?
    \item[D4.] \label{check:D4} Was the data collection protocol approved (or determined exempt) by an ethics review board?
    \item[D5.] \label{check:D5} Did you report the basic demographic and geographic characteristics of the annotator population that is the source of the data?
\end{enumerate}

\subsection*{E. Did you use AI assistants (e.g., ChatGPT, Copilot) in your research, coding, or writing?}
\label{check:E}
\begin{enumerate}
    \item[E1.] \label{check:E1} If you used AI assistants, did you include information about their use?
\end{enumerate}

\end{tcolorbox}

\section{\texttt{NO} Justification Examples}

Authors need to give missing information, or an explanation. Underneath is examples of both, where $P=$ \texttt{YES}, $S=$ \texttt{NO}.
\begin{tcolorbox}[
    label=box:no-responses,
    enhanced,
    colback=white, 
    colframe=gray!80, 
    arc=2pt, 
    title=\textbf{\texttt{NO} Justification Examples}, 
    colbacktitle=gray!80,
    fonttitle=\bfseries,
    attach boxed title to top left={xshift=3mm, yshift=-3mm},
    boxed title style={colframe=gray!80, colback=gray!80}
]
\footnotesize
\medskip
\textbf{(a) Explanation} \\
\textit{Q: Did you discuss the license or terms for use and/or distribution of any artifacts?} \\
\textbf{J:} The artifacts are used with their original license. 

\medskip
\textbf{(b) Missing information} \\
\textit{Q: If you used AI assistants, did you include information about their use?} \\
\textbf{J:} We only used AI tools (e.g. Grammarly) for small grammatical corrections.

\end{tcolorbox}

\section{Normalisation Algorithm Diagram}

\begin{figure}[htbp]
  \centering
  \begin{tikzpicture}[
      >=Stealth,
      font = \small\sffamily,
      every node/.style = {align=center},
      leaf/.style = {draw, rounded corners, minimum width=2cm, minimum height=0.7cm, inner sep=4pt},
      llmbox/.style = {draw, rounded corners, minimum width=3cm, minimum height=1.3cm, inner sep=4pt, fill=blue!5}
    ]

    \node[leaf, fill=gray!10] (pdf) at (0,0) {PDF (Checklist)};

    \node[leaf] (md) at (-1.5,-2.5) {Markdown};
    \node[leaf] (jpeg) at (1.5,-2.5) {JPEG};

    \node[llmbox] (llm) at (0,-4.5) {LLM};

    \node[leaf] (json) at (0,-6.5) {Structured JSON};

    \draw[->] (pdf.south) -- (md.north)
        node[midway, left, font=\footnotesize] {Marker};
    \draw[->] (pdf.south) -- (jpeg.north)
        node[midway, right, font=\footnotesize] {PyMuPDF};

    \coordinate (llmInLeft)  at ($(llm.north)+(-1.0,0)$);
    \coordinate (llmInRight) at ($(llm.north)+(1.0,0)$);
    \draw[->] (md.south)  -- (llmInLeft);
    \draw[->] (jpeg.south) -- (llmInRight);

    \draw[->] (llm.south) -- (json.north);

  \end{tikzpicture}
  \caption{Simplified checklist processing pipeline. The LLM also has an input of the structured JSON Schema, with all 23 questions.}
  \label{fig:simple}
\end{figure}
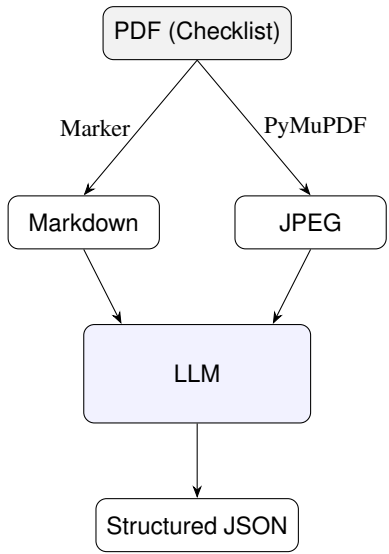  

\section{Main Analysis}
\subsection{Parent-Child Table}

\begin{table}[htbp]
\centering
\footnotesize
\label{tab:main-inconsistencies}
\begin{tabular}{@{}lrr@{}}
\toprule
Section & Inconsistent Papers & Rate (\%) \\ \midrule
A       & 0                   & 0.0       \\
B       & 171                 & 9.5       \\
C       & 50                  & 2.8       \\
D       & 77                  & 4.3       \\
E       & 12                  & 0.7       \\ \midrule
Overall & 281                 & 15.5      \\
\bottomrule
\end{tabular}
\caption{Parent-child logical inconsistencies across the 1,809 main-track papers.}
\end{table} 

\section{Comparison: Main \& Findings}
In all analysis, the main track is cyan whilst the findings track is indigo.

\subsection{\texttt{NO} Justification Quality}
Comparison between both tracks. Recall empty is 0 words, whilst brief is <10 words. 

\begin{figure}[htbp]
    \centering
    \includegraphics[width=\columnwidth]{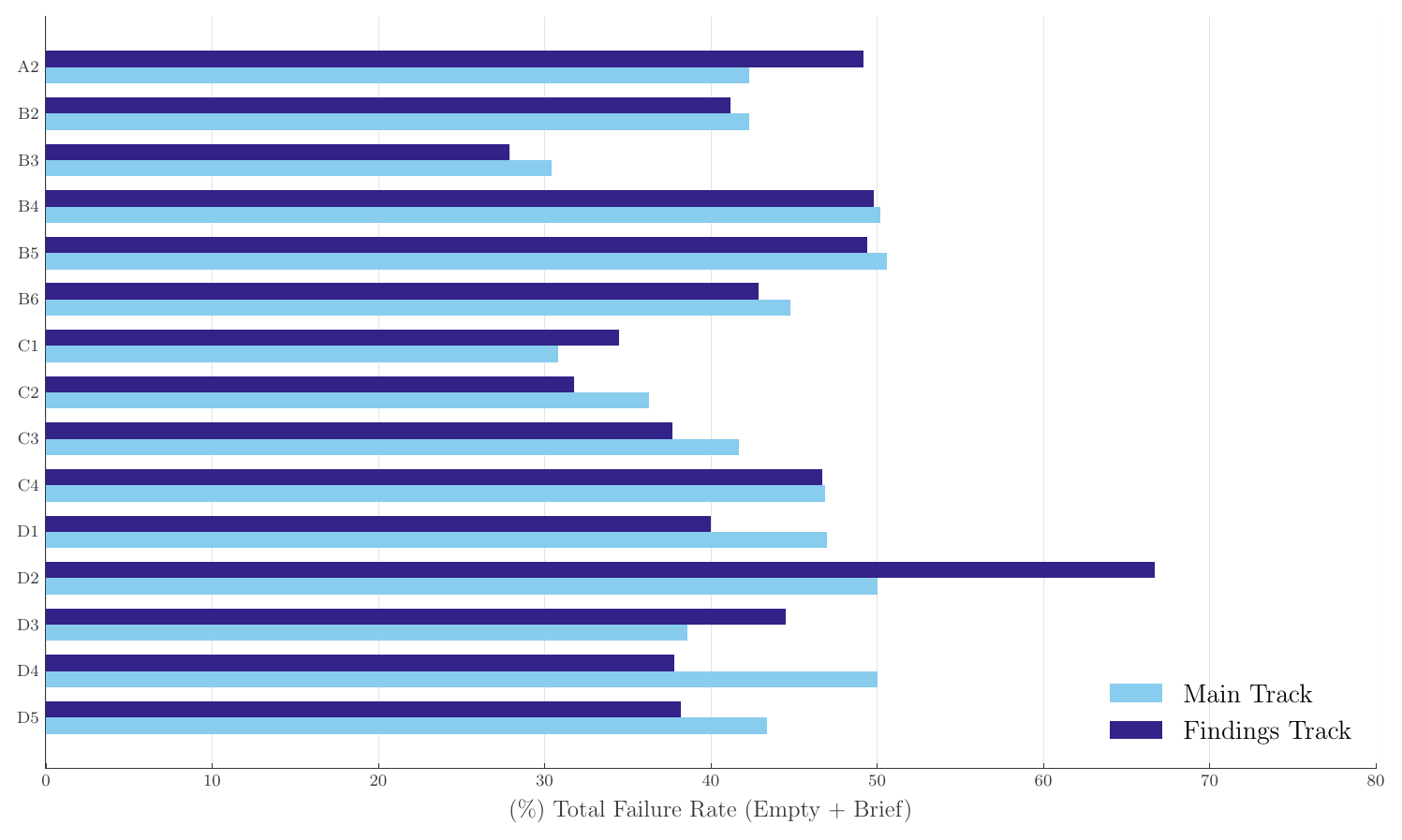}
    \caption{Justification failure rates (empty + brief) by subquestion, grouped by track..}
    \label{fig:comp-justification}
\end{figure}

\subsection{Comparison Findings-Main Answers}
\begin{figure}[htbp]
\centering
\includegraphics[width=\columnwidth]{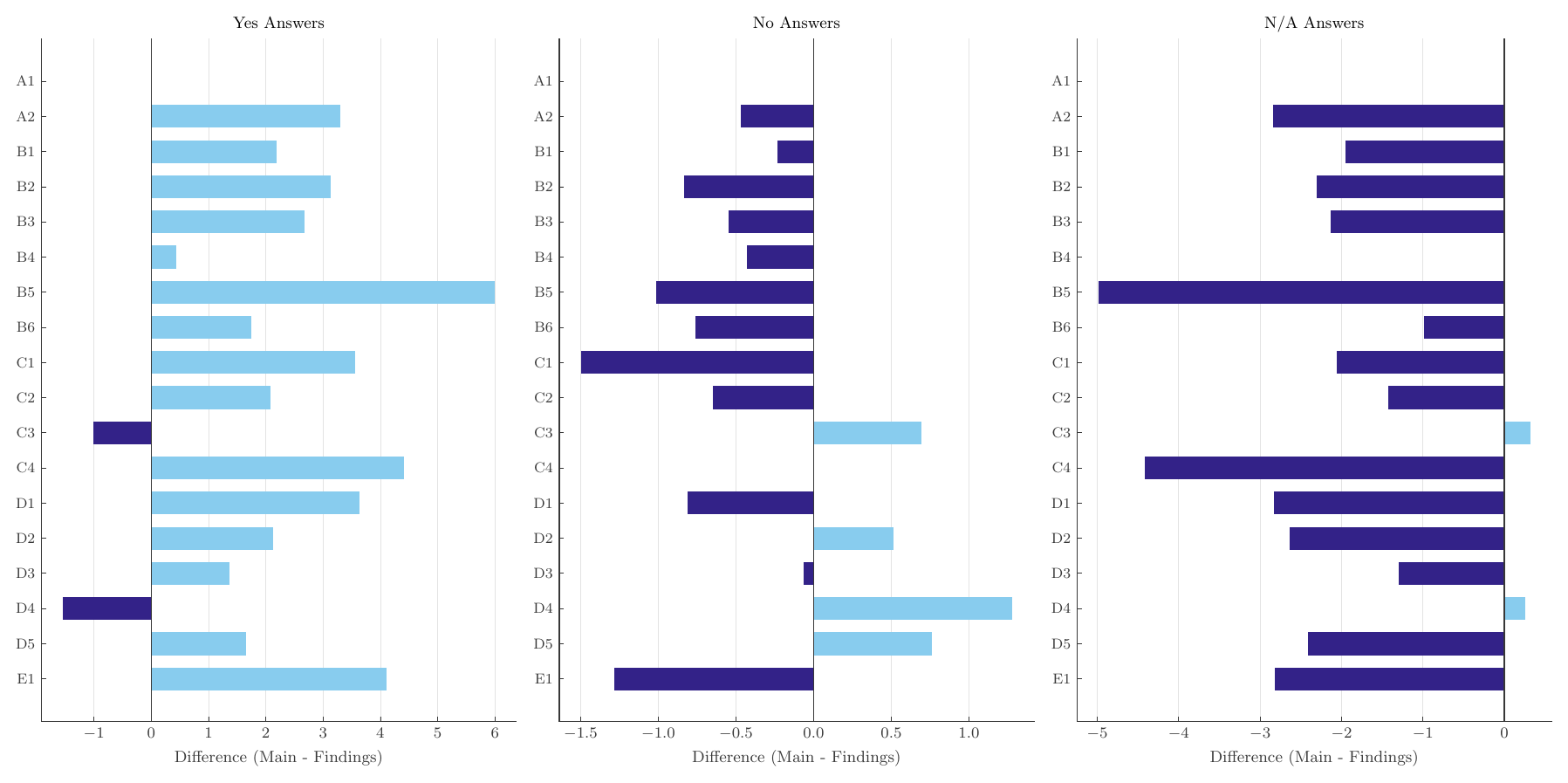}
\caption{Percentage-point differences (Main minus Findings) for Yes, No, and \texttt{N/A} responses across all 18 subquestions.}
\label{fig:comp-answers}
\end{figure}

\begin{figure}[htbp]
\centering
\includegraphics[width=\columnwidth]{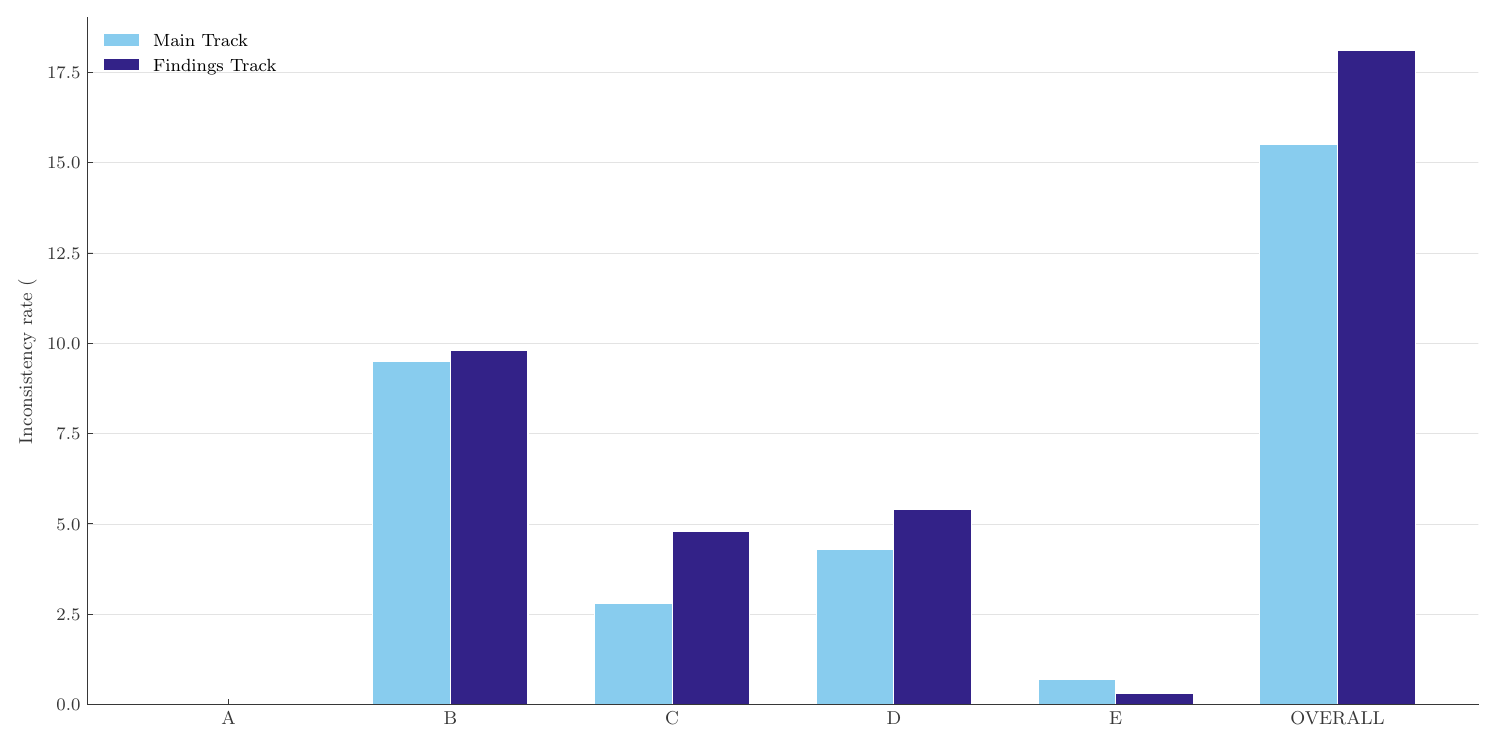}
\caption{Parent-child inconsistency rates by section, grouped by track.}
\label{fig:comp-inconsistency}
\end{figure}

\end{document}